\documentclass[11pt, a4paper, onecolumn, copyright, gr]{google}

\usepackage[authoryear, sort&compress, round]{natbib}
\uselogo{} 

\title{TabFM: A Zero-Shot Foundation Model for Tabular Data}

\correspondingauthor{weihaokong@google.com}

\renewcommand{\today}{2026-09-29}

\author[1]{Weihao Kong}
\author[1]{Erez Louidor Ilan}
\author[1]{Shuxin Nie}
\author[1]{Taman Narayan}
\author[1]{Rajat Sen}
\author[1]{Yichen Zhou}
\author[1]{Deqing Fu}
\author[1]{Samet Oymak}
\author[1]{Abhimanyu Das}

\affil[1]{Google Research}

\usepackage{amsmath,amsfonts,bm}

\def\1{\bm{1}}

\DeclareMathAlphabet{\mathsfit}{\encodingdefault}{\sfdefault}{m}{sl}
\SetMathAlphabet{\mathsfit}{bold}{\encodingdefault}{\sfdefault}{bx}{n}

\usepackage{subcaption}

\graphicspath{{figures/}}
\definecolor{tabfmblue}{HTML}{3186FF}
\definecolor{fmblue}{HTML}{1A73E8}
\newcommand{\tabfm}{\mbox{\texttt{TabFM}}}
\newcommand{\tabfmp}{\mbox{\texttt{TabFM+}}}
\newcommand{\tabfmauto}{\mbox{\texttt{TabFM-Auto}}}
\newcommand{\best}[1]{\textbf{#1}}
\renewcommand{\copyrightext}{\footerfont Model Weights: \url{https://huggingface.co/google/tabfm-1.1.0-pytorch} \\ \footerfont \textcopyright\, \the\year{} Google. All rights reserved}

\begin{abstract}
Tabular machine learning typically relies on per-dataset workflows, fitting tree ensembles or running AutoML searches from scratch for every task. We present \tabfm{}, a 400M-parameter tabular foundation model that formulates supervised tabular prediction as in-context learning. \tabfm{} produces calibrated zero-shot predictions in a single forward pass without task-specific tuning. Trained entirely on synthetic tables generated from structural causal models, \tabfm{} learns general tabular representations that transfer zero-shot to real-world tasks. Across all 51 benchmark datasets in TabArena (38 classification and 13 regression), zero-shot \tabfm{} ranks first among default tabular foundation models and outperforms tuned AutoML pipelines. Two extensions over the same frozen weights improve performance further on both tracks: multi-view feature expansion with ensembling and post-hoc calibration (\tabfmp{}), and LLM-guided, dataset-specific data processing and feature engineering (\tabfmauto{}).
\end{abstract}

\begin{document}

\maketitle

\begin{figure*}[!h]
\centering
\includegraphics[width=\linewidth]{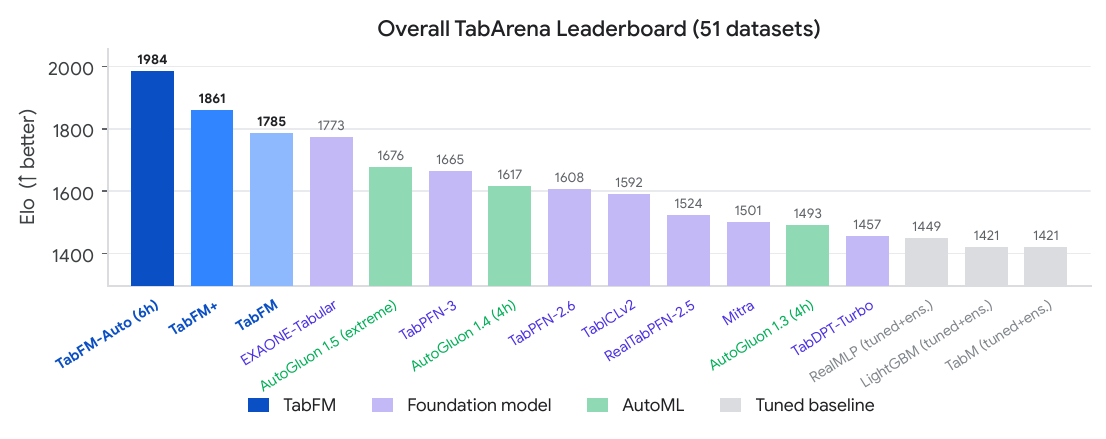}
\caption{Overall TabArena Elo ratings for the top 16 methods across all 51 benchmark datasets (38 classification and 13 regression). \tabfm{} ranks first among zero-shot tabular foundation models, while \tabfmp{}'s multi-view feature expansion, ensembling, and calibration and \tabfmauto{}'s LLM-guided feature engineering with \texttt{Gemini-3.8-Flash} take the top two positions overall without updating any weights.}
\label{fig:headline}
\end{figure*}

\section{Introduction} \label{sec:intro}

\begin{figure}[t]
\centering
\includegraphics[width=\linewidth]{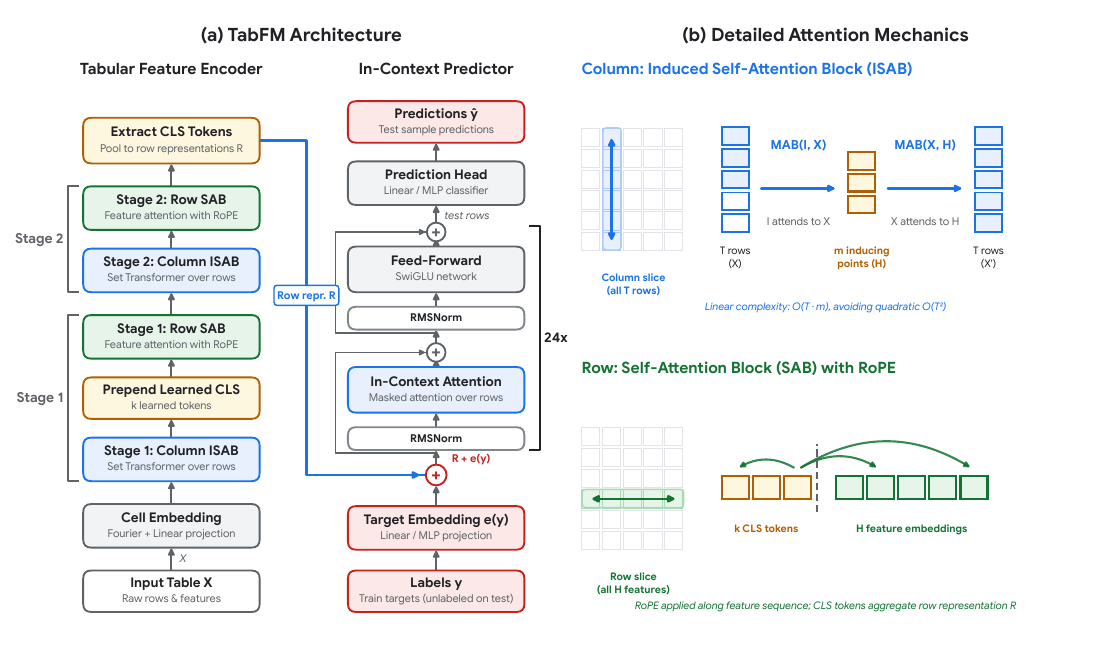}
\caption{The \tabfm{} Architecture. (a) Tables are embedded cell-wise ($T \times H \times E$), compressed by two untied stages alternating column ISAB and row SAB with 8 CLS tokens into row representations $\mathbf{R}$, conditioned on $\mathbf{e}(y)$ for train rows, then processed by a 24-layer in-context predictor. (b) Column ISAB attends across rows in linear time, while row SAB attends across features with RoPE.}
\label{fig:arch}
\end{figure}

Supervised learning on tabular data relies primarily on tree ensembles, most prominently gradient-boosted decision trees (GBDTs)~\citep{chen2016xgboost,ke2017lightgbm,prokhorenkova2018catboost} and Random Forests~\citep{breiman2001randomforests}, alongside AutoML systems~\citep{feurer2015autosklearn,erickson2020autogluon} that stack heterogeneous models. Tree ensembles are effective because axis-aligned splits reliably handle mixed column types, missing entries, and irregular feature scales. However, these methods follow a \emph{per-dataset optimization} paradigm: each new task demands custom preprocessing, split-finding sweeps, and hyperparameter search from scratch, so every dataset is learned from scratch.

Foundation models instead amortize learning into a single forward pass. Following Prior-Data Fitted Networks (PFNs)~\citep{muller2022pfn,hollmann2023tabpfn}, an in-context tabular model parametrizes a universal posterior predictive map $f_{\boldsymbol{\theta}}(\mathcal{D}_{\text{train}}, \mathbf{x}_{\text{test}}) \approx p(y_{\text{test}} \mid \mathbf{x}_{\text{test}}, \mathcal{D}_{\text{train}})$ over labeled context $\mathcal{D}_{\text{train}} = \{(\mathbf{x}_i, y_i)\}_{i=1}^{T_{\text{train}}}$ and unlabeled queries. Theoretically, transformers realize implicit optimization algorithms in their forward activations~\citep{vonoswald2023transformers,garg2022what,li2023transformers,fu2024second}. Scaling this approach to general tables is constrained by the structure of tabular data. Rows are exchangeable, requiring permutation equivariance, while columns mix continuous values that require fine numerical resolution with discrete categories that lack natural ordering. Naive attention over all cells scales as $\mathcal{O}(T^2 H^2)$ for $T$ rows and $H$ columns, historically confining tabular transformers to fewer than 1,000 rows~\citep{hollmann2023tabpfn}.

We introduce \tabfm{}, a 400M-parameter Transformer that addresses these constraints (Figure~\ref{fig:arch}). At the input layer, dyadic feature grouping pairs each column with neighbors at offsets $0, 1, 3$ and projects cell values through learned Fourier frequency banks. \tabfm{} decouples feature encoding from cross-instance reasoning by alternating linear-time column-wise Induced Self-Attention Blocks (ISAB)~\citep{lee2019settransformer} with row-wise Self-Attention Blocks (SAB) using Rotary Position Embeddings (RoPE)~\citep{su2021roformer}. Learned CLS tokens pool arbitrary feature counts into fixed-width row representations, scaling context to 16,384 instances, and an asymmetric mask enforces conditional independence among queries.

We evaluate three deployment tiers on TabArena~\citep{erickson2025tabarena}. Zero-shot \tabfm{} ranks first among default foundation models on both tracks. With the weights kept frozen, \tabfmp{} (see \S\ref{sec:ensemble}) combines cross and SVD feature expansion with $K=32$ multi-view ensembling, stacked through regularized Non-Negative Least Squares~\citep{lawson1995solving} and Platt calibration~\citep{platt1999probabilistic}, and \tabfmauto{} (see \S\ref{sec:tabfm_auto} and \citealt{fu2026tabfmauto}) pairs the same frozen model with Gemini 3.8 Flash in a closed program-synthesis loop over dataset-specific feature engineering, taking the top rank on both suites.

\section{Related Work} \label{sec:related_work}

\paragraph{Tabular Foundation Models and In-Context Optimization.}
In-context tabular prediction was introduced by Prior-Data Fitted Networks (PFNs)~\citep{muller2022pfn}, which train networks on synthetic priors to approximate Bayesian posterior predictives. TabPFN~\citep{hollmann2023tabpfn} demonstrated in-context classification on small tables, followed by extensions to regression and larger samples~\citep{hollmann2025tabpfn2,grinsztajn2025tabpfn25,grinsztajn2026tabpfn3}, continued pretraining on empirical corpora~\citep{garg2025realtabpfn,ma2024tabdpt,hosseinzadeh2026tabdptturbo}, and decoupled row-column architectures with inducing points~\citep{qu2025tabicl,qu2026tabiclv2}. Theoretically, in-context learning is well described as implicit algorithmic optimization: attention layers execute iterative update steps in their activations~\citep{vonoswald2023transformers,garg2022what,li2023transformers}, and deeper stacks attain second-order Newton convergence on ill-conditioned regression~\citep{fu2024second}. \tabfm{} builds on this design by pairing an alternating column- and row-attention encoder with a deep 24-layer in-context predictor over spectral cell embeddings.

\paragraph{Tree Ensembles, Deep Architectures, and Numerical Embeddings.}
GBDTs~\citep{chen2016xgboost,ke2017lightgbm,prokhorenkova2018catboost} build axis-aligned boundaries that are robust to uninformative features and invariant to monotone rescaling, and benchmarks consistently show they outperform per-dataset deep models~\citep{shwartz2022tabular,grinsztajn2022why,mcelfresh2023when} such as TabNet~\citep{arik2021tabnet}, FT-Transformer~\citep{gorishniy2021revisiting}, SAINT~\citep{somepalli2021saint}, and TabM~\citep{gorishniy2024tabm}. Standard neural layers also struggle to encode continuous scalars without losing numerical resolution. Transformers learn Fourier representations when grokking arithmetic~\citep{nanda2023progress}, language models build internal Fourier features for numerical scale~\citep{zhou2024pretrained} that recur across architectures~\citep{fu2026convergent}, and Fourier Number Embeddings preserve numeric resolution~\citep{zhou2025fone}. \tabfm{} adopts learned Fourier cell embeddings on this basis.

\paragraph{AutoML and Programmatic Data Optimization.}
AutoML automates model selection, feature engineering, and tuning~\citep{feurer2015autosklearn}. AutoGluon-Tabular~\citep{erickson2020autogluon} combines multi-layer stacking and bagging over a per-task search budget. Language models have also been applied to tabular transformation~\citep{tornede2024automl}, through row serialization in TabLLM~\citep{hegselmann2023tabllm} and feature-engineering synthesis in CAAFE~\citep{hollmann2023caafe}. \tabfmauto{}~\citep{fu2026tabfmauto} instead wraps a frozen tabular foundation model in a closed cross-validation loop, synthesizing dataset-specific feature engineering pipelines without updating model weights.

\section{TabFM: Model Architecture and Pretraining} \label{sec:arch_and_data}

\subsection{Model Architecture}
\label{sec:arch}

\tabfm{} factorizes tabular prediction into four stages: spectral cell embedding, alternating column-wise and row-wise attention, CLS pooling, and deep in-context prediction. Figure~\ref{fig:arch} shows the pipeline and Table~\ref{tab:arch_params} its specification.

\begin{table}[t]
\centering
\caption{\tabfm{} architecture specification (400M checkpoint, $E=d_{\text{model}}=128$). All sublayers use sandwich RMSNorm with SwiGLU activations, zero dropout, and no bias parameters.}
\label{tab:arch_params}
\small
\setlength{\tabcolsep}{4pt}
\renewcommand{\arraystretch}{1.15}
\begin{tabularx}{\linewidth}{@{} l l cccc r >{\raggedright\arraybackslash}X @{}}
\toprule
& & \multicolumn{4}{c}{\textbf{Configuration}} & & \\
\cmidrule(lr){3-6}
\textbf{Stage} & \textbf{Module} & Width & Heads & Induc. & FFN & \textbf{Params} & \textbf{Function} \\
\midrule
Cell Embedder      & Spectral proj.        & 128  & --       & --  & --   & 0.05\,M  & Learned Fourier features \\
Column Embedder    & $2\times3$ ISAB       & 128  & 4        & 256 & 512  & 3.3\,M   & Linear row attention \\
Row Interaction    & $2\times3$ SAB        & 128  & 8        & --  & 512  & 1.6\,M   & RoPE feature attention \\
ICL Predictor      & $24\times$ SAB        & 1024 & 8        & --  & 4096 & 402.7\,M & Masked in-context mixing \\
Prediction Head    & 2-layer MLP           & 1024 & --       & --  & 1024 & 1.05\,M  & Output \\
\midrule
\rowcolor{tabfmblue!10}
\textbf{\tabfm{}}  & \multicolumn{5}{l}{} & \textbf{408.7\,M} &  \\
\bottomrule
\end{tabularx}
\end{table}


\paragraph{Cell Embedder and Feature Grouping.}
Given an input table $\mathbf{X} \in \mathbb{R}^{T \times H}$ with $T = T_{\text{train}} + T_{\text{test}}$ rows $\mathbf{x}_i \in \mathbb{R}^H$, \tabfm{} applies \emph{feature grouping} following TabPFN-3~\citep{grinsztajn2026tabpfn3} and TabICLv2~\citep{qu2026tabiclv2} to capture local cross-feature interactions before attention. Each column $j \in \{0, \dots, H-1\}$ is grouped with columns at dyadic offsets $j_g = (j + 2^g - 1) \bmod H$ for $g \in \{0, 1, 2\}$ (group size $G = 3$, offsets $0, 1, 3$). Each slot $g$ projects its scalar $v$ through a slot-specific bank of 32 learned Fourier frequencies $\boldsymbol{\omega}_g \in \mathbb{R}^{32}$~\citep{tancik2020fourier,zhou2025fone}:
\begin{equation}
  \boldsymbol{\gamma}_g(v) = \big[\cos(2\pi \omega_{g,1} v),\, \sin(2\pi \omega_{g,1} v),\, \dots,\, \cos(2\pi \omega_{g,32} v),\, \sin(2\pi \omega_{g,32} v)\big]^\top \in \mathbb{R}^{64}.
\end{equation}
Numerical and categorical columns route per slot through separate frequency banks ($\boldsymbol{\omega}_g^{\mathrm{num}}$ vs.\ $\boldsymbol{\omega}_g^{\mathrm{cat}}$) and projections ($\mathbf{W}_g^{\mathrm{num}}, \mathbf{W}_g^{\mathrm{cat}} \in \mathbb{R}^{64 \times E}$), keeping continuous values on a metric scale while mapping categories to distinct embeddings. Summing across slots gives $\mathbf{X}^{(0)}_{i,j} = \sum_{g=0}^{G-1} \boldsymbol{\gamma}_g(x_{i, j_g})\,\mathbf{W}_g \in \mathbb{R}^E$ ($E = 128$). Labeled rows receive a target embedding $\mathbf{e}(y_i)$ added to each cell, while query rows receive nothing.

\paragraph{Attention Foundations and Column Embedding.}
To process both dimensions of a table without quadratic cell attention, column-wise attention treats the $T$ rows as the sequence within each column, capturing marginal distributions, whereas row-wise attention treats the $H$ features as the sequence within each row, capturing cross-feature structure.

We first fix notation. For queries $\mathbf{Q} \in \mathbb{R}^{N_q \times d_k}$, keys $\mathbf{K} \in \mathbb{R}^{N_{kv} \times d_k}$, values $\mathbf{V} \in \mathbb{R}^{N_{kv} \times d_v}$, and an additive mask $\mathbf{M} \in \{0, -\infty\}^{N_q \times N_{kv}}$ that deletes forbidden query--key pairs, scaled dot-product attention and its $h$-head form over $\mathbf{X} \in \mathbb{R}^{N_q \times d}$, $\mathbf{Y} \in \mathbb{R}^{N_{kv} \times d}$ (with $d_k = d_v = d/h$) are
\begin{equation}
  \mathrm{Attn}(\mathbf{Q}, \mathbf{K}, \mathbf{V};\, \mathbf{M}) = \mathrm{softmax}\!\left(\frac{\mathbf{Q}\mathbf{K}^\top}{\sqrt{d_k}} + \mathbf{M}\right)\!\mathbf{V},
  \quad
  \mathrm{MHA}(\mathbf{X}, \mathbf{Y};\, \mathbf{M}) = \big[\mathrm{head}_1, \dots, \mathrm{head}_h\big]\mathbf{W}^O,
\end{equation}
with $\mathrm{head}_i = \mathrm{Attn}\big(\mathbf{X}\mathbf{W}_i^Q,\, \mathbf{Y}\mathbf{W}_i^K,\, \mathbf{Y}\mathbf{W}_i^V;\, \mathbf{M}\big)$ and learnable $\mathbf{W}_i^Q, \mathbf{W}_i^K \in \mathbb{R}^{d \times d_k}$, $\mathbf{W}_i^V \in \mathbb{R}^{d \times d_v}$, $\mathbf{W}^O \in \mathbb{R}^{d \times d}$. Normalization uses $\mathrm{RMSNorm}$~\citep{zhang2019rmsnorm} and the feed-forward sublayer uses $\mathrm{SwiGLU}$~\citep{shazeer2020glu}:
\begin{equation}
  \mathrm{RMSNorm}(\mathbf{x}) = \frac{\mathbf{x}}{\sqrt{\tfrac{1}{d}\sum_{j=1}^d x_j^2 + \epsilon}} \odot \mathbf{g},
  \qquad
  \mathrm{SwiGLU}(\mathbf{x}) = \Big(\mathrm{swish}(\mathbf{x}\mathbf{W}_1) \odot (\mathbf{x}\mathbf{V}_1)\Big)\mathbf{W}_2,
\end{equation}
where $\mathbf{g} \in \mathbb{R}^d$ is a learnable gain, $\epsilon = 10^{-6}$, $\mathrm{swish}(z) = z\,\sigma(z)$, and $\mathbf{W}_1, \mathbf{V}_1 \in \mathbb{R}^{d \times 4d}$, $\mathbf{W}_2 \in \mathbb{R}^{4d \times d}$.

\tabfm{} uses two fixed attention masks. Attention along the feature axis carries a padding mask $\mathbf{M}^{\mathrm{pad}}$ that masks the unused slots of the $H_{\max}$-column buffer into which a table of $H \le H_{\max}$ columns is padded ($H_{\max} = 100$ throughout pretraining, \S\ref{sec:pretrain}), and attention along the row axis carries a context mask $\mathbf{M}^{\mathrm{ctx}}$ that admits only the labeled prefix as keys:
\begin{equation}
  M^{\mathrm{pad}}_{jl} = \begin{cases} 0 & l \le H \\ -\infty & \text{otherwise,} \end{cases}
  \qquad
  M^{\mathrm{ctx}}_{ik} = \begin{cases} 0 & k \le T_{\text{train}} \\ -\infty & \text{otherwise.} \end{cases}
\end{equation}
$\mathbf{M}^{\mathrm{ctx}}$ does not depend on the query index $i$: both training and test queries attend to the same key set $\{1, \dots, T_{\text{train}}\}$. Writing $\mathbf{0}$ for the all-zero mask, an unmasked block is the special case $\mathbf{M} = \mathbf{0}$.

The Set Transformer Multihead Attention Block ($\mathrm{MAB}$)~\citep{lee2019settransformer} composes these with sandwich normalization, applying $\mathrm{RMSNorm}$ at both the input and output of each sublayer while residuals carry unnormalized activations:
\begin{align}
  \mathbf{Z} &= \mathbf{X} + \mathrm{RMSNorm}\Big(\mathrm{MHA}\big(\mathrm{RMSNorm}(\mathbf{X}),\, \mathrm{RMSNorm}(\mathbf{Y});\, \mathbf{M}\big)\Big), \\
  \mathrm{MAB}(\mathbf{X}, \mathbf{Y};\, \mathbf{M}) &= \mathbf{Z} + \mathrm{RMSNorm}\Big(\mathrm{SwiGLU}\big(\mathrm{RMSNorm}(\mathbf{Z})\big)\Big).
\end{align}
Self-attention and induced self-attention follow directly, with $\mathbf{I} \in \mathbb{R}^{m \times E}$ a set of $m = 256$ learned inducing points:
\begin{equation}
  \mathrm{SAB}(\mathbf{X};\, \mathbf{M}) = \mathrm{MAB}(\mathbf{X}, \mathbf{X};\, \mathbf{M}),
  \qquad
  \mathrm{ISAB}_m(\mathbf{X};\, \mathbf{M}) = \mathrm{MAB}\Big(\mathbf{X},\, \mathrm{MAB}(\mathbf{I}, \mathbf{X};\, \mathbf{M});\, \mathbf{0}\Big).
\end{equation}
The inner block compresses $T$ instances onto $m$ inducing vectors and the outer block reads them back, reducing $\mathcal{O}(T^2 E)$ attention to $\mathcal{O}(TmE)$. Masking the inner projection with $\mathbf{M}^{\mathrm{ctx}}$ ensures that the inducing summaries $\mathrm{MAB}(\mathbf{I}, \mathbf{X};\, \mathbf{M}^{\mathrm{ctx}})$ are computed solely from the labeled training rows $k \le T_{\text{train}}$, so the unmasked outer read-back distributes training-set column statistics to all rows without leaking information across test instances. Column-wise attention is therefore $\mathrm{ISAB}_{256}(\cdot\,;\, \mathbf{M}^{\mathrm{ctx}})$ over the row axis.

\paragraph{Row Interaction and Context Pooling.}
Row-wise attention runs across features within each row via $\mathrm{SAB}(\cdot\,;\, \mathbf{M}^{\mathrm{pad}})$ (8 heads, width $E=128$), with Rotary Position Embeddings (RoPE)~\citep{su2021roformer} modulating queries and keys along the feature axis by rotations $\mathbf{R}_{\Theta, p}$. Confining RoPE to features distinguishes channels without imposing ordinality and leaves rows permutation-equivariant. Feature positions are therefore encoded relatively rather than through a learned table, so $H$ is not fixed by the weights and inference can run wider than pretraining (\S\ref{sec:ensemble}). \tabfm{} alternates two untied stages of three $\mathrm{ISAB}_{256}(\cdot\,;\, \mathbf{M}^{\mathrm{ctx}})$ blocks and three $\mathrm{SAB}(\cdot\,;\, \mathbf{M}^{\mathrm{pad}})$ blocks. Eight learned CLS tokens $\mathbf{C} \in \mathbb{R}^{8 \times E}$ are prepended along the feature axis after the first column stage, and their final states are concatenated into $\mathbf{r}_i \in \mathbb{R}^{1024}$, giving $\mathbf{R} \in \mathbb{R}^{T \times 1024}$.

\paragraph{In-Context Learning Predictor.}
The predictor reads the row sequence $\mathbf{R}$. Labeled rows are conditioned on their target a second time, $\tilde{\mathbf{r}}_i = \mathbf{r}_i + \mathbf{e}(y_i)$ for $i \le T_{\text{train}}$, while the target embedding of a query row is zeroed before it is added. The predictor stacks 24 $\mathrm{SAB}(\cdot\,;\, \mathbf{M}^{\mathrm{ctx}})$ layers (width 1024, 8 heads, FFN 4096, 402.7M parameters), supplying the depth that multi-step implicit optimization requires~\citep{li2023transformers,fu2024second}. Because $\mathbf{M}^{\mathrm{ctx}}$ masks all keys with index $k > T_{\text{train}}$, each test row $i > T_{\text{train}}$ attends exclusively to the labeled context, and its own features propagate to the prediction through the query projection and the residual stream. Test predictions are therefore conditionally independent given $\mathcal{D}_{\text{train}}$, which prevents label leakage and lets the query block be split or reordered without changing predictions. A sandwich-normalized two-layer MLP head then maps $\tilde{\mathbf{r}}_i$ to logits $\hat{\mathbf{y}}_i \in \mathbb{R}^C$ or a scalar $\hat{y}_i$.

\paragraph{Attention Complexity.}
Decoupling feature encoding from cross-row prediction reduces both time and memory complexity. Column-wise attention over $T$ rows is $\mathcal{O}(THmE)$ with $m = 256$ inducing points rather than the $\mathcal{O}(T^2HE)$ of full cross-row attention over cells, and row-wise attention is $\mathcal{O}(TH^2E)$ along the feature axis, which stays bounded because the feature count is far below the row count. Only the predictor attends across all $T$ rows at full width, at $\mathcal{O}(T^2d)$ with $d = 1024$, and it does so once on pooled row representations rather than once per cell. Context therefore reaches 16,384 instances, an order of magnitude beyond the regime in which in-context tabular prediction was first demonstrated~\citep{hollmann2023tabpfn}.

\subsection{Synthetic Pretraining Data and Curriculum}
\label{sec:pretrain}

\begin{figure}[t]
\centering
\includegraphics[width=\linewidth]{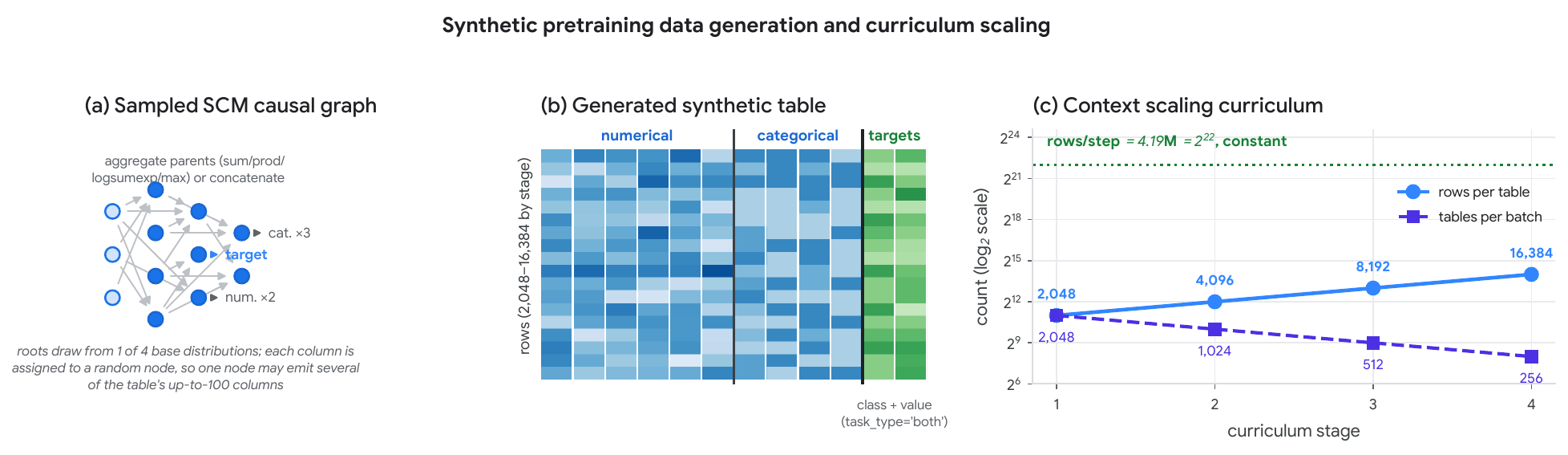}
\caption{Synthetic pretraining and curriculum: (a) a sampled SCM graph with randomized functional dependencies, (b) the resulting table with numerical, categorical, and dual targets, and (c) the four-stage curriculum scaling context from 2,048 to 16,384 instances at constant tokens per step.}
\label{fig:datagen}
\end{figure}

\tabfm{} is pretrained entirely on synthetic tables drawn from structural causal models (SCMs)~\citep{pearl2009causality}, which generate structured feature dependencies (Figure~\ref{fig:datagen}). Each dataset samples a directed acyclic graph with randomized functional dependencies~\citep{qu2026tabiclv2}. Root variables propagate through non-linear transformations and algebraic aggregations to produce tables mixing continuous and discrete columns.

To match the distributional variations of real-world datasets, each sample jointly randomizes table shape, the share of categorical columns and their cardinalities, missing entries, label noise, and class balance, with the feature axis capped at 100 columns. Randomizing these jointly with the graph prevents the model from overfitting to a fixed table geometry and keeps the evaluation tables of Section~\ref{sec:eval} inside the support of the pretraining distribution. Every sampled table is split into a labeled context and a query block before it reaches the model, so the pretraining objective matches the inference-time interface: the loss is evaluated only on query rows, jointly over the classification and the regression target that each table carries.

\paragraph{Curriculum Learning.}
A four-stage curriculum grows the context from 2,048 to 16,384 instances (Figure~\ref{fig:datagen}(c)). At each transition rows per table double while batch size halves, holding tokens per optimization step fixed at $2^{22} \approx 4.19$M. Keeping per-step compute invariant lets the model adapt to longer contexts without destabilizing optimization. Training on shorter contexts with large batch sizes first stabilizes the Fourier cell embedder and the alternating column and row attention stages, after which the longer-context stages adapt the inducing-point bottlenecks and 24-layer predictor to larger sample sizes.

\subsection{TabArena Zero-Shot Evaluation}
\label{sec:eval}

We evaluate on TabArena~\citep{erickson2025tabarena}, a benchmark of 51 datasets (38 classification, 13 regression), each evaluated under repeated 10-fold cross-validation. Performance across heterogeneous metrics is summarized by Bradley--Terry Elo~\citep{elo1978rating,bradley1952rank,hunter2004mm}, which places per-dataset evaluation metrics on a common scale.

\paragraph{Protocol and Metrics.}
The comparison pool holds 67 method configurations including tuned tree ensembles, tuned deep baselines, AutoML systems at four-hour budgets, and published tabular foundation models. Ratings come from TabArena's Bradley--Terry fit over pairwise outcomes at the level of a single (dataset, fold) pair, weighting every dataset equally, anchoring Random Forest at 1000, and taking the median rating over 100 bootstrap rounds. Alongside Elo, we report three summary metrics across datasets. G-mean is the geometric mean over datasets of the mean error, which weights relative improvements equally across datasets of different difficulty. Wins sums, over datasets, the fraction of folds a method takes outright with ties split. Improvability averages $1 - \mathrm{err}_{\text{best}} / \mathrm{err}_{\text{method}}$ over datasets, measuring the relative error reduction achievable by a per-dataset oracle selector.

\begin{table}[t]
\centering
\caption{TabArena benchmark leaderboards under repeated 10-fold cross-validation. Ratings are obtained from a unified Bradley--Terry fit anchored to RF (default) at 1000 Elo. G-mean is the geometric mean error across datasets, Wins counts fold-averaged dataset victories, and Improvability is the relative error reduction an oracle selector would achieve.}
\label{tab:leaderboards}
\vspace{2pt}
\begin{subtable}[t]{0.495\linewidth}
\centering
\caption{Classification (38 datasets, 594 tasks)}
\label{tab:cls}
\scriptsize
\setlength{\tabcolsep}{2.0pt}
\begin{tabular}{clcccc}
\toprule
\# & Method & Elo $\uparrow$ & Wins $\uparrow$ & Improv.\ $\downarrow$ & gmean $\downarrow$ \\
\midrule
\rowcolor{tabfmblue!12}
\best{1} & \best{TabFM-Auto} & \best{1940.7} & \best{13.50} & \best{2.47\%} & \best{0.0921}\\
\rowcolor{tabfmblue!12}
\best{2} & \best{TabFM+} & \best{1838.0} & \best{6.63} & \best{5.22\%} & \best{0.0948}\\
\rowcolor{tabfmblue!12}
\best{3} & \best{TabFM} & \best{1768.6} & \best{5.54} & \best{6.10\%} & \best{0.0957}\\
4 & EXAONE-Tabular & 1768.0 & 3.00 & 9.47\% & 0.1009\\
5 & AutoGluon 1.5 (ext.) & 1669.7 & 1.48 & 10.00\% & 0.1010\\
6 & TabPFN-3 & 1641.9 & 0.40 & 12.34\% & 0.1050\\
7 & AutoGluon 1.4 (4h) & 1621.9 & 0.26 & 13.33\% & 0.1079\\
8 & TabPFN-2.6 & 1590.3 & 0.03 & 13.87\% & 0.1075\\
9 & TabICLv2 & 1586.4 & 0.57 & 13.08\% & 0.1060\\
10 & RealTabPFN-2.5 & 1536.0 & 0.07 & 14.22\% & 0.1077\\
11 & AutoGluon 1.3 (4h) & 1475.6 & 0.14 & 16.27\% & 0.1147\\
12 & RealMLP (tuned) & 1436.7 & 0.08 & 17.34\% & 0.1161\\
\bottomrule
\end{tabular}
\end{subtable}
\hfill
\begin{subtable}[t]{0.495\linewidth}
\centering
\caption{Regression (13 datasets, 222 tasks)}
\label{tab:reg}
\scriptsize
\setlength{\tabcolsep}{2.0pt}
\begin{tabular}{clcccc}
\toprule
\# & Method & Elo $\uparrow$ & Wins $\uparrow$ & Improv.\ $\downarrow$ & gmean $\downarrow$ \\
\midrule
\rowcolor{tabfmblue!12}
\best{1} & \best{TabFM-Auto} & \best{2392.4} & \best{9.61} & \best{0.00\%} & \best{15.96}\\
\rowcolor{tabfmblue!12}
\best{2} & \best{TabFM+} & \best{2189.2} & \best{1.09} & \best{1.34\%} & \best{16.17}\\
\rowcolor{tabfmblue!12}
\best{3} & \best{TabFM} & \best{2055.2} & \best{0.92} & \best{2.62\%} & \best{16.40}\\
4 & EXAONE-Tabular & 1973.1 & 0.18 & 3.64\% & 16.58\\
5 & TabPFN-3 & 1866.6 & 0.17 & 3.33\% & 16.51\\
6 & AutoGluon 1.5 (ext.) & 1851.2 & 0.14 & 4.83\% & 16.78\\
7 & TabPFN-2.6 & 1791.4 & 0.00 & 4.91\% & 16.79\\
8 & AutoGluon 1.4 (4h) & 1732.2 & 0.03 & 5.57\% & 16.91\\
9 & TabICLv2 & 1723.6 & 0.28 & 4.85\% & 16.78\\
10 & TabDPT-Turbo & 1660.8 & 0.14 & 6.07\% & 17.02\\
11 & AutoGluon 1.3 (4h) & 1646.8 & 0.03 & 7.14\% & 17.23\\
12 & RealMLP (tuned) & 1623.4 & 0.03 & 6.54\% & 17.09\\
\bottomrule
\end{tabular}
\end{subtable}
\end{table}

\begin{figure}[t]
\centering
\includegraphics[width=\linewidth]{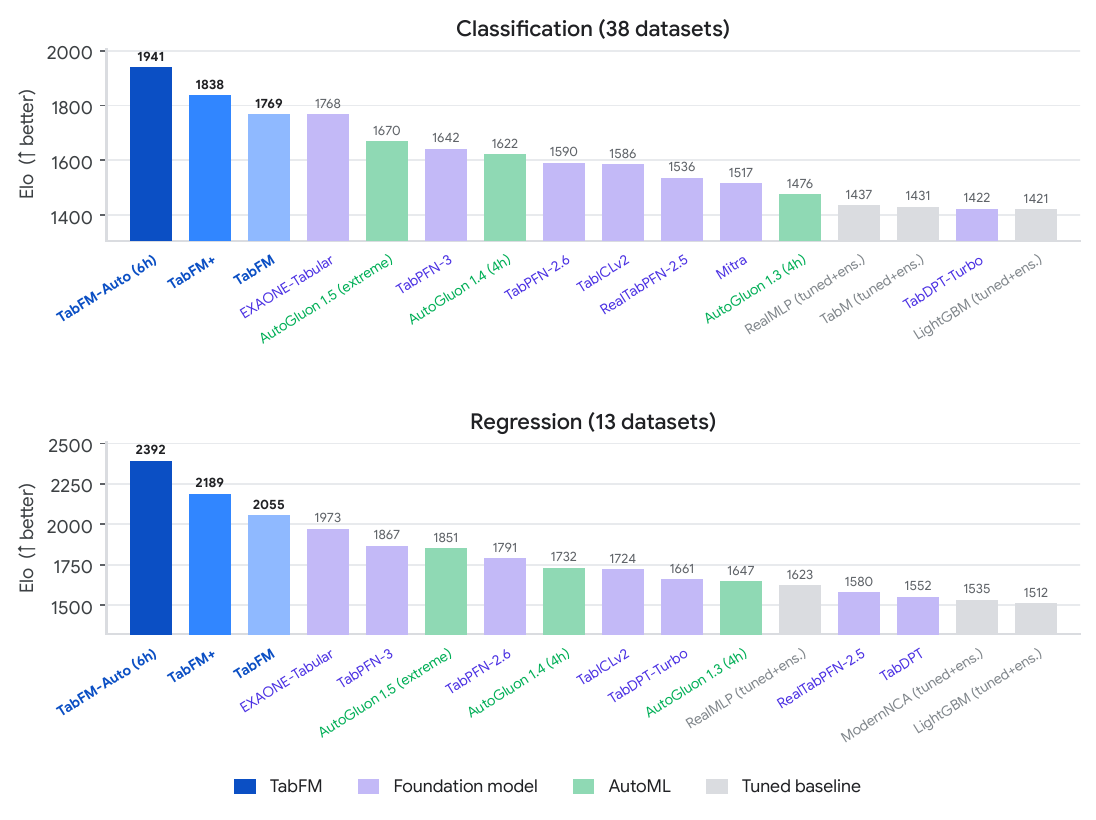}
\caption{Task-separated TabArena Elo over 38 classification (top) and 13 regression (bottom) datasets, top 16 methods per suite. Colors mark \tabfm{} variants in blue, peer foundation models in purple, AutoML in green, and tuned baselines in grey.}
\label{fig:elo:separated}
\end{figure}

\paragraph{Zero-Shot Performance.}
\tabfm{} leads both tracks among default models (Table~\ref{tab:leaderboards}, Figure~\ref{fig:elo:separated}). On regression it reaches 2055.2~Elo, ahead of EXAONE-Tabular (1973.1), TabPFN-3 (1866.6), AutoGluon~1.5 extreme (1851.2), and TabICLv2 (1723.6), with the lowest geometric-mean error (16.40) and 2.62\% oracle improvability among single-pass models. On classification it reaches 1768.6~Elo, ahead of EXAONE-Tabular (1768.0), AutoGluon~1.5 extreme (1669.7), TabPFN-3 (1641.9), and TabICLv2 (1586.4), again with the lowest geometric-mean error (0.0957), the most outright wins (5.54), and the lowest oracle improvability (6.10\%). The regression margin is the larger of the two, consistent with spectral embeddings that place targets on a continuous scale rather than on piecewise-constant splits. On classification, where the top four methods fall within a 130-Elo span, \tabfm{}'s advantage comes from lower regret across datasets, reducing oracle improvability from 9.47\% (EXAONE-Tabular) and 10.00\% (AutoGluon~1.5 extreme) to 6.10\%.

\begin{figure}[t]
\centering
\includegraphics[width=\linewidth]{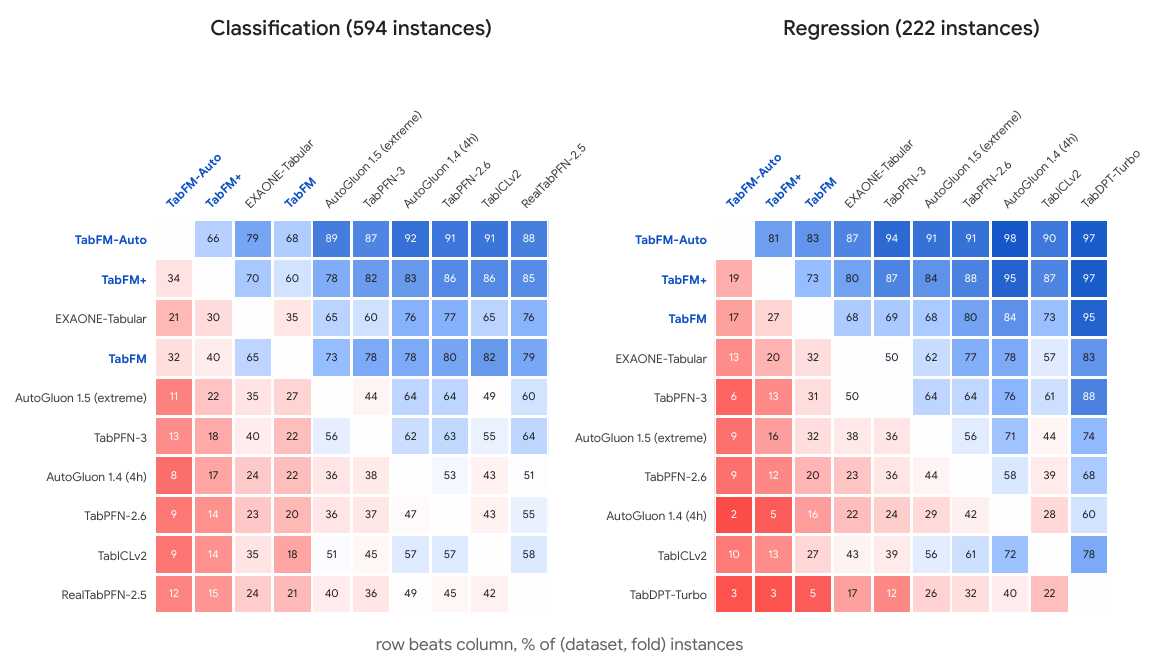}
\caption{Pairwise win rates on TabArena: percentage of dataset-fold instances where the row method beats the column method, ties counted as half. Methods are ordered by Elo.}
\label{fig:winrate}
\end{figure}

Pooled over both suites (Figure~\ref{fig:headline}), \tabfm{} ranks first among default foundation models at 1785.4~Elo and wins most head-to-head folds against every baseline (Figure~\ref{fig:winrate}): 65.7\% against EXAONE-Tabular, 72.1\% against AutoGluon~1.5 extreme, 75.9\% against TabPFN-3, and 79.8\% against TabICLv2. Win rates increase monotonically with Elo difference across tuned GBDTs, deep tabular models, 4-hour AutoML ensembles, and peer foundation models.

\section{TabFM+: Feature Engineering and Inference-Time Ensembling} \label{sec:ensemble}

Because \tabfm{} applies dyadic feature grouping and $\mathrm{RoPE}$ along the column axis, its forward pass is invariant to row order but sensitive to column order, numerical scaling, and appended feature interactions. \tabfmp{} uses this property at test time by running $K = 32$ transformed views of the input table through the frozen checkpoint. After standardizing raw columns (expanding datetimes into five numerical channels, merging categories with frequency below two, and dropping constant or duplicate columns), \tabfmp{} constructs two pools of engineered features for an $H$-column table: \emph{multiplicative cross features}, which sample up to $k_{\mathrm{cross}} = \lfloor\sqrt{H}\rfloor$ numerical column pairs $(c_1, c_2)$ uniformly at random from all $\binom{H_{\mathrm{num}}}{2}$ pairs and append products $x_{i,c_1} \cdot x_{i,c_2}$, and \emph{Truncated SVD structural features}, which one-hot encode categoricals alongside standardized numericals and extract up to $k_{\mathrm{svd}} = \lfloor\sqrt{H}\rfloor$ leading singular vectors. Crosses inject pairwise non-linearities, while SVD components supply dense low-rank summaries of global linear structure.

To diversify the $K = 32$ forward passes, \tabfmp{} splits the member budget: half of the members evaluate the unaugmented original columns (capped at 500 features, well beyond the 100 columns seen during pretraining), while the other half append random draws from the cross and SVD pools. Each member then applies an independent view transformation combining alternative preconditioning (standard scaling with either identity or Yeo--Johnson power normalization and $z$-score outlier clipping at $4.0$), random column permutations, random bijective categorical index permutations, and cyclic shifts of classification labels. Because both dyadic feature grouping and row RoPE depend on column order, permuting columns alters both the input-layer feature triplets and their relative positional encodings, producing distinct internal views from one set of weights. Member predictions $\hat{\mathbf{y}}^{(1)}, \dots, \hat{\mathbf{y}}^{(K)}$ are stacked via Non-Negative Least Squares~\citep{lawson1995solving} weights $\mathbf{w} \in \Delta^{K-1}$ fit on internal validation folds and shrunk toward uniform as $\mathbf{w}_{\text{final}} = 0.75\,\mathbf{w} + 0.25\,\tfrac{1}{K}\mathbf{1}$, with Platt scaling~\citep{platt1999probabilistic} for classification. This adds $+69.4$~Elo on classification and $+134.0$~Elo on regression (second overall among 67 methods, Table~\ref{tab:leaderboards}), with regression gaining nearly twice as much because averaging reduces variance more on continuous scales than on sharp class posteriors.

\section{TabFM-Auto: Feature Engineering with Gemini} \label{sec:tabfm_auto}

While \tabfmp{} applies generic perturbations to the input table, many datasets depend on domain-specific features. \tabfmauto{} pairs the frozen \tabfm{} checkpoint with \texttt{Gemini-3.8-Flash}, which is given a dataset and iteratively writes, evaluates, and refines a Python pipeline around the pretrained model. The program specifies data preprocessing, feature engineering, the selection of rows that enter the context, post-processing of the prediction, and the model's inference-time settings. \tabfm{} remains the primary predictor with its weights frozen: auxiliary models may be fitted and blended, but \tabfm{} parameters are never updated. We summarize the method here and refer to \citet{fu2026tabfmauto} for full details.

\begin{figure}[t]
\centering
\includegraphics[width=\linewidth]{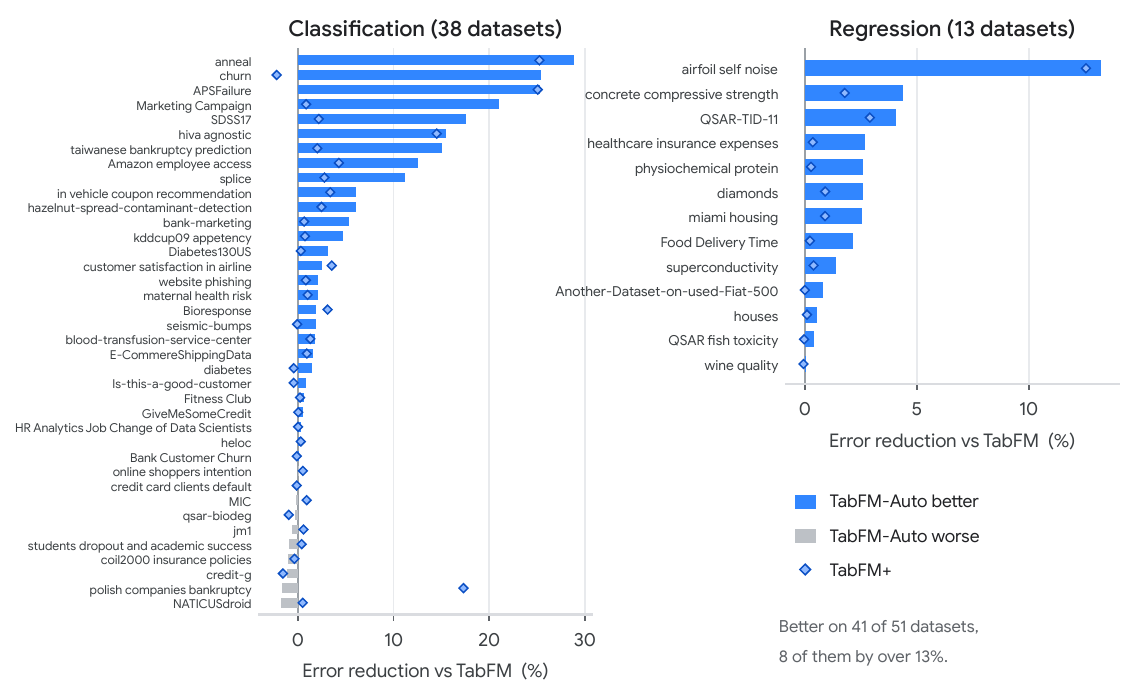}
\caption{Per-dataset relative error reduction ($1 - \mathrm{err}_{\mathrm{method}}/\mathrm{err}_{\mathrm{TabFM}}$) over zero-shot \tabfm{} across all 51 TabArena datasets. Bars show \tabfmauto{} and diamonds show \tabfmp{}. \tabfmauto{} improves 41 of 51 datasets, ten by more than 10\%.}
\label{fig:autogains}
\end{figure}

Each candidate program is scored by three-fold cross-validation inside the training split of the first fold with 8 ensemble members, and its score, its per-fold values, and any traceback are appended to a log that is read before the next edit. Candidate programs execute in a sandbox without access to held-out test splits or evaluation folds. Every run starts from the identity program (vanilla \tabfm{}), so that any improvement comes from the synthesized transformations around zero-shot \tabfm{}. Search stops after 96 evaluations or six hours, whichever comes first. Gemini then selects one program from its top-3 validation candidates, and that program is refit on every published fold and scored once on test rows.

Improvements come primarily from feature construction rather than hyperparameter tuning, for example Strouhal and Reynolds numbers on \texttt{airfoil\_self\_noise} or ICD-9 diagnoses folded into chapters on \texttt{Diabetes130US}. Splitting the final programs by whether any hook changed the table, the median error reduction over zero-shot \tabfm{} is 2.14\% when it did and 0.27\% when it did not.

\tabfmauto{} takes the top position on both TabArena classification and regression tasks, gaining $+172.1$ and $+337.2$ Elo over the base \tabfm{} (see Table~\ref{tab:leaderboards}). It improves 41 of the 51 datasets and beats inference-time ensembling on 40 of them (Figure~\ref{fig:autogains}). Regression improves on all 13 datasets and reaches 0.00\% oracle improvability. On the ten classification datasets that do not improve over zero-shot \tabfm{}, test error increases by at most 1.8\%.

\section{Conclusion} \label{sec:conclusion}

We presented \tabfm{}, a 400M-parameter Transformer that performs supervised tabular prediction through in-context learning. Trained exclusively on synthetic tables generated from structural causal models, \tabfm{} transfers zero-shot to real-world tasks and ranks first among default foundation models on TabArena, while inference-time ensembling (\tabfmp{}) and LLM-guided feature engineering (\tabfmauto{}) bring further gains without updating any weights.

Current pretraining covers synthetic numerical and categorical tables of at most 16,384 rows and 100 columns, leaving larger tables to length generalization or subsampling and encoding free text without semantic tokenization. Scaling pretraining directly to million-row and thousand-column tables is an immediate next step, using hierarchical row pooling and sparse feature attention to keep those shapes tractable. Combining synthetic causal priors with real-world tabular corpora and pretrained text and temporal encoders would likewise extend in-context learning to multimodal tables with free-text fields, timestamps, and high-cardinality identifiers, and generalizing the architecture from single flat tables to relational schemas would avoid manually joining and flattening databases before inference. More broadly, whereas \tabfmauto{} queries \tabfm{} as a fixed evaluator inside an outer LLM search loop, distilling synthesized feature pipelines back into pretraining could fold feature discovery directly into the forward pass.

\bibliography{main}

\end{document}